\pdfoutput=1

\documentclass[11pt]{article}

\usepackage{booktabs}   
\usepackage{array}      
\usepackage{caption}
\usepackage[final]{acl}

\usepackage{times}
\usepackage{latexsym}

\usepackage[T1]{fontenc}

\usepackage[utf8]{inputenc}

\usepackage{microtype}

\usepackage{inconsolata}

\usepackage{graphicx}
\usepackage{makecell}
\usepackage{amssymb}
\usepackage{multirow}
\usepackage{enumitem}
\usepackage[font=small,skip=5pt]{caption}
\title{Tabular Data Understanding with LLMs: \\ A Survey of Recent Advances and Challenges}

\author{
  Xiaofeng Wu  \quad Alan Ritter \quad Wei Xu  \\
  College of Computing, Georgia Institute of Technology \\
  \texttt{xwu414@gatech.edu},  \texttt{alan.ritter@cc.gatech.edu}, \texttt{wei.xu@cc.gatech.edu}
}

\begin{document}
\maketitle
\begin{abstract}

Tables have gained significant attention in large language models (LLMs) and multimodal large language models (MLLMs) due to their complex and flexible structure. Unlike linear text inputs, tables are two-dimensional, encompassing formats that range from well-structured database tables to complex, multi-layered spreadsheets, each with different purposes. This diversity in format and purpose has led to the development of specialized methods and tasks, instead of universal approaches, making navigation of table understanding tasks challenging.
To address these challenges, this paper introduces key concepts through a taxonomy of tabular input representations and an introduction of table understanding tasks. We highlight several critical gaps in the field that indicate the need for further research: (1) the predominance of retrieval-focused tasks that require minimal reasoning beyond mathematical and logical operations; (2) significant challenges faced by models when processing complex table structures, large-scale tables, length context, or multi-table scenarios; and (3) the limited generalization of models across different tabular representations and formats.

\end{abstract}

\section{Introduction}
\begin{figure}
    \centering
    \includegraphics[width=1\linewidth]{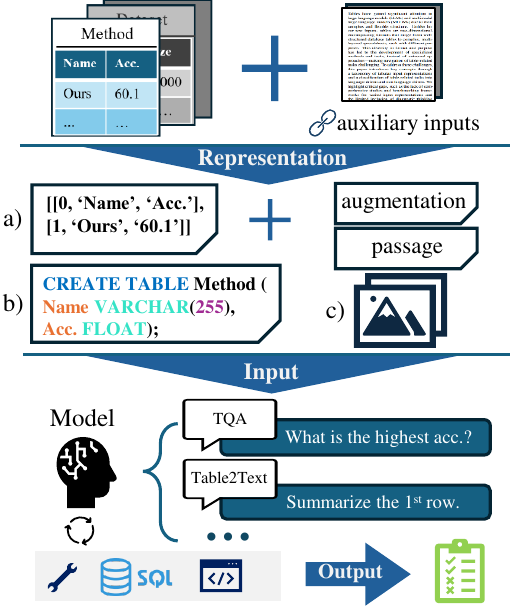}
    \caption{Workflow of table-related tasks in large models. Tables or databases, possibly accompanied by additional input data, are transformed into input representations, which could take the form of (a) serialization, (b) database schema, (c) images, or other format with optional augmentations. These inputs are then processed by models usually leveraging SQL, and other tools to generate task specific outputs.}
    \vspace{-10pt}
    \label{fig:tablePipe}
\end{figure}

Tables have garnered increasing attention due to advances in large language models (LLMs) and multi-modal large language models (MLLMs), owing to the unique challenges they present. Unlike linear text, tabular data possess an inherently visual, two-dimensional format that requires specialized pipelines to be processed effectively, as shown in Figure \ref{fig:tablePipe}. Additionally, tables exhibit structural flexibility, serving a wide range of purposes—from well-structured database tables to hierarchical, multi-layered spreadsheets and multimedia-linked info-boxes. These variations in purpose and structure have driven the development of diverse input representations, tasks, and specialized methods and datasets. However, such specialization often comes at the expense of universality \cite{tablellama}, making it difficult for new researchers to navigate the field effectively.  
While existing surveys \cite{survey1, survey2, survey3, survey4, survey5} have explored various prompting, training, and transformer-based methods for table processing, there is a need for a comprehensive survey that uncovers new opportunities, focusing on tasks and benchmarks in tabular understanding.

\begin{figure*}[th]
    \centering
    \includegraphics[width=1\linewidth]{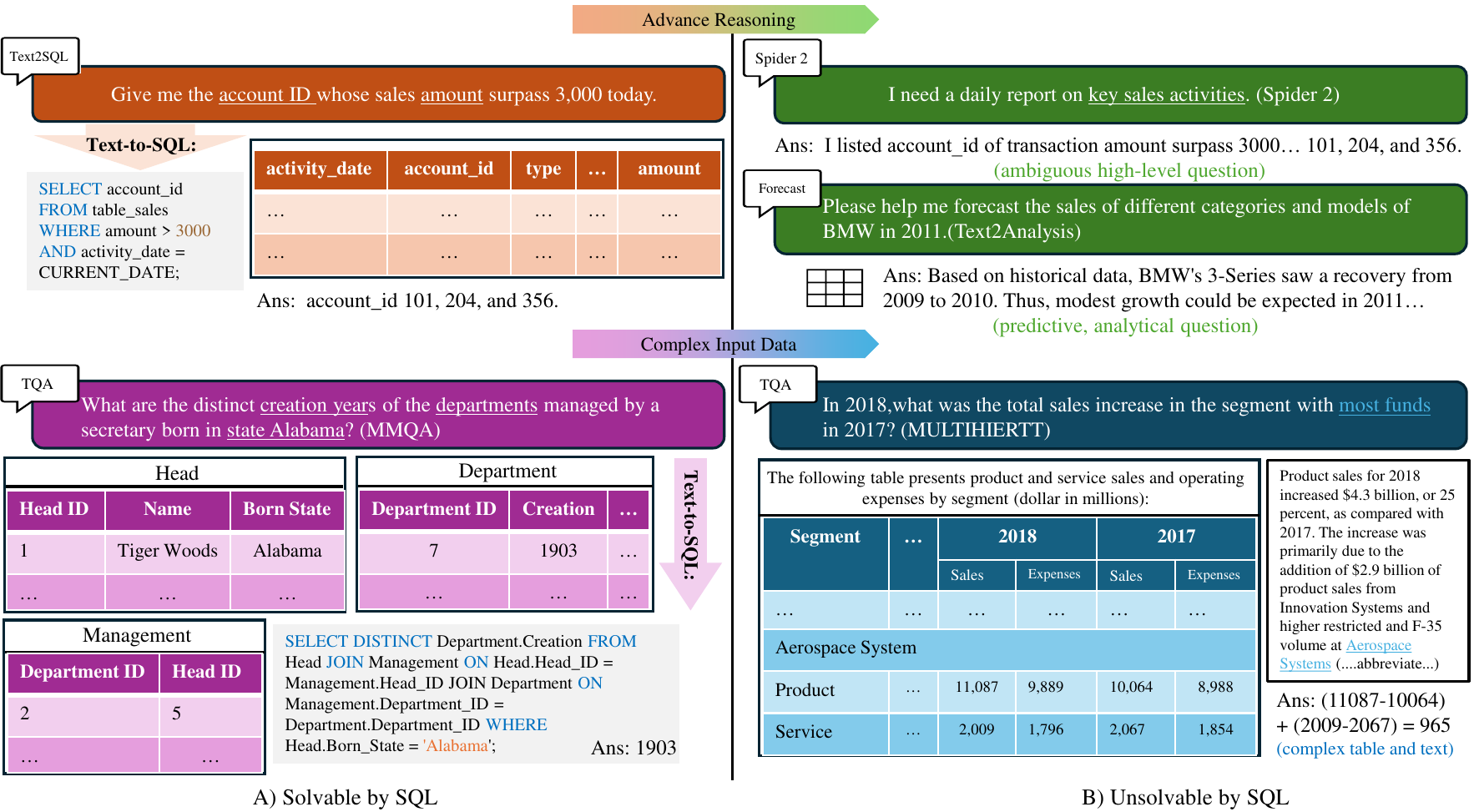}
   \caption{The left side illustrates examples of tasks that can be addressed with SQL-based methods such as typical Text-to-SQL task and a Table QA task from MMQA \cite{mmqa2}. In contrast, the right side presents tasks that demand advanced reasoning or involve complex inputs, such as those found in Spider 2 \cite{spider2}, Text2Analysis \cite{text2analysis}, and MULTIHIERTT \cite{multihiertt}, which go beyond the capabilities of SQL-based approaches.}
    \label{fig:task2sql}
\end{figure*}


To address the existing gap and assist researchers in navigating table-related tasks, this paper presents a systematic taxonomy of tabular data representations and introduces a broad range of both well-established and novel tasks. For instance, we examine \textit{Table QA}, which focuses on answering natural language questions based on table content, and \textit{Table-to-Text}, which involves generating natural language summaries from tabular data. We also highlight innovative tasks such as \textit{leaderboard construction}, which aggregates result tables from scientific papers to provide a comprehensive comparison of methods in one specific field. For well-established tasks, we compile key benchmarks and their associated table formats, categorizing improvements in newer benchmarks relative to earlier ones to highlight emerging research trends.


Furthermore, our survey reveals new opportunities by focusing on tasks and challenges identified in widely used benchmarks. Despite significant progress in prompting and training methods—as highlighted in existing surveys \cite{survey3, survey4, survey5}—and the robust performance of recent tabular foundational models that integrate tabular data during the pre-training and fine-tuning stages of 72B base models \cite{tablegpt2}, current table processing benchmarks tend to concentrate on limited reasoning tasks and often rely on simplistic, synthetic tables with inconsistent input representations. While effective for initial evaluations, these benchmarks fall short in assessing the performance of more advanced methods and models in real-world scenarios that require higher-level reasoning and the processing of complex inputs, ultimately limiting their generalizability and broader applicability.


\section{Findings and Future Direction}

In this section, we outline three key findings that underscore the need for further investigation. 

\subsection{Limited Scope Beyond Mathematical Reasoning}
Recent work has begun to saturate performance on many widely used benchmarks. For example, question-decomposition pipelines have yielded significant improvements \cite{dailqsql, dater, chainoftable}; the method proposed by \citet{artemisda} achieved over 80\% accuracy on the Wiki-Table Questions benchmark \cite{WTQ} and more than 93\% on TabFact \cite{tabFact}, two popular datasets for table QA and fact verification. Moreover, the success of table foundation models—integrating specialized table encoders into large-scale language models pre-trained and fine-tuned on tabular data \cite{tablegpt2}—signals a growing trend toward applying tabular methods to larger models. These advances suggest it is time to move beyond data retrieval-based tasks, as most benchmarks rely on detailed queries that prompt models to extract specific information from tables using logical operations.

Many existing benchmarks are even constructed by first generating SQL queries or sequences of mathematical expressions, which are then translated into natural language query \cite{WTQ, SQA, multiTabQA, mmqa2}, or by framing questions whose answers can be fully derived using mathematical functions \cite{imTQA, crtQA, multihiertt, openWikiTable}. While efforts have focused on enhancing task complexity through additional reasoning steps or embedding complex mathematical functions, the core structure of these tasks remains fundamentally unchanged. As shown in Figure \ref{fig:task2sql}, such descriptive questions can be solved relatively easily by text-to-SQL methods when tables are well-structured.

\begin{figure}[t]
    \centering
    \includegraphics[width=1\linewidth]{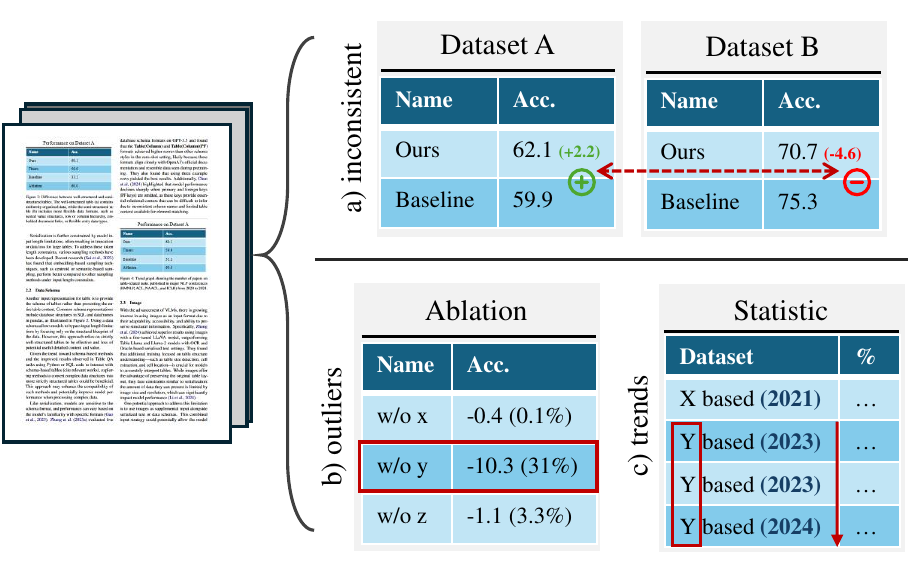}
    \caption{Illustration of the proposed task: Scientific Document Understanding with Tables which require diagnosing implicit knowledge embedded in tabular data, which may not be well addressed in text. Examples include: a) inconsistent results under conditions; b) outliers in values; and c) key trends.}
    \vspace{-5pt}
    \label{fig:docUnderstanding}
\end{figure}

Notably, recent work \cite{discoveryBench} has further pushed the boundaries by emphasizing higher-order reasoning skills. For example, \citet{text2analysis} introduced tasks that extend beyond basic descriptive analysis, such as insight identification, similar to what is shown in Figure \ref{fig:docUnderstanding}, which demands diagnostic thinking; forecasting, which requires predictive thinking; and chart creation from ambiguous queries, a task that requires prescriptive thinking—selecting the appropriate chart type and determining optimal intervals to produce visually appealing figures. In these tasks, models cannot simply rely on finding synonyms or related attributes in the table to perform data retrieval. Instead, they must understand the overall context of the table and the user’s intent to address the query.

A similar direction is explored by Spider 2 \cite{spider2}, which introduces questions requiring higher levels of reasoning. Unlike benchmarks such as Spider \cite{spider} and its extensions, which introduce marginal difficulties by swapping explicit schema names with synonyms or rephrasing utterances \cite{spiderRealistic, spiderSYN}, Spider 2 presents high-level, intent-driven queries, as illustrated in Figure \ref{fig:task2sql}. For example, instead of asking explicitly (e.g., ``\textit{Give me the account ID whose sales surpass a threshold today}''), Spider 2 poses abstract, goal-oriented queries (e.g., ``\textit{I need a daily report on key sales activities}''). These queries challenge models to infer the user’s intent, requiring a deep understanding of both the database schema and the query’s broader context. Furthermore, \citet{practiq} introduce multi‑turn conversations that teach models to seek clarification whenever a user’s initial query is ambiguous, thereby better mirroring real‑world interactions and mitigating the multiple‑interpretation issue identified by \citet{problemsInSpider}.
\vspace{-1pt}

\begin{figure*}[ht]
    \centering
    \includegraphics[width=1\linewidth]{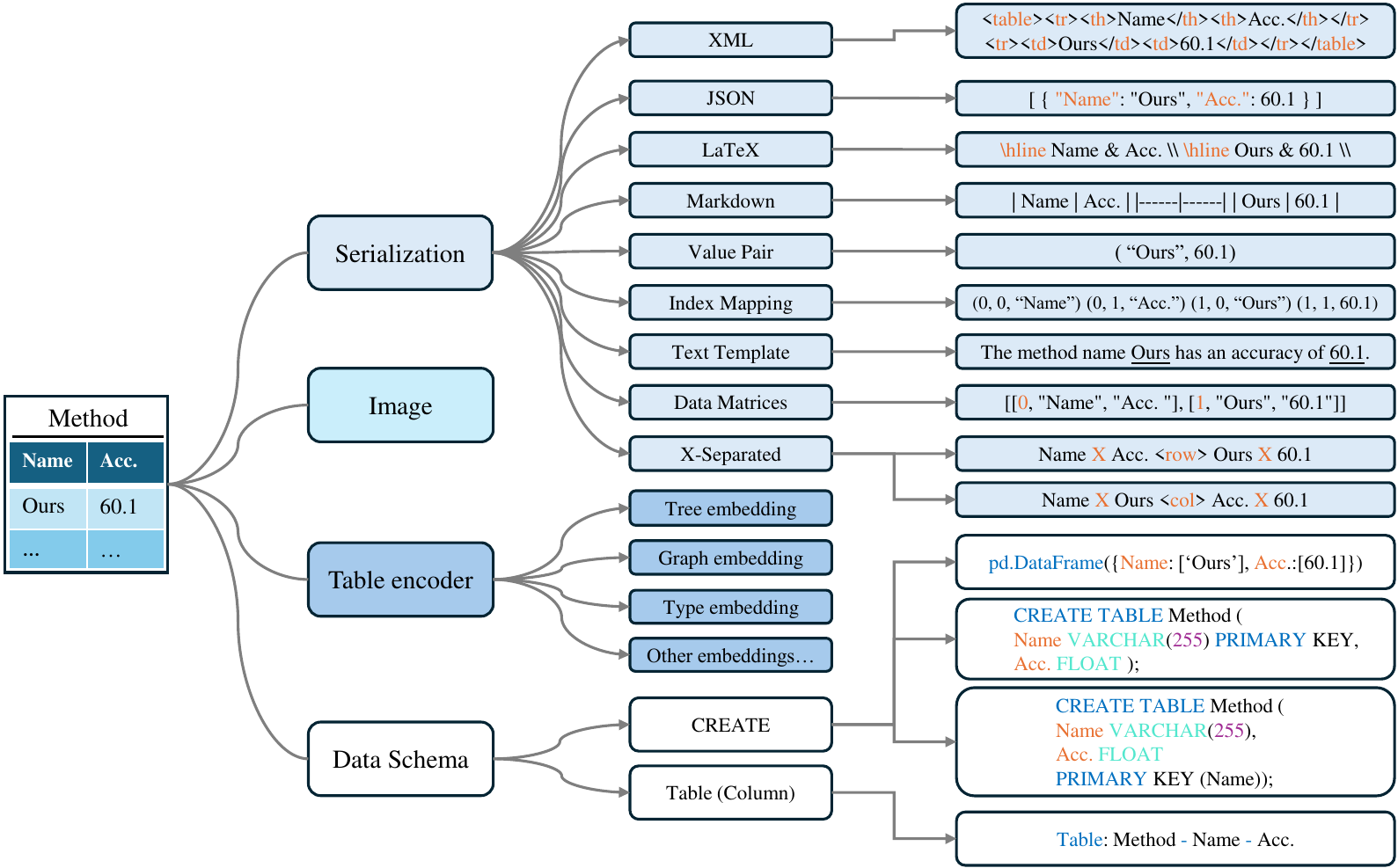}
    \caption{Taxonomy of table input representation methods, encompassing serialization, image, specialized table encoders, and data schema. Examples illustrating each representation type are shown on the right.}
    \vspace{-10pt}
    \label{fig:inputRepresentation}
\end{figure*}

\subsection{Lack of Robustness on Input Complexity}

Another area of opportunity in current table-related research is enhancing model robustness when processing complex input scenarios, including intricate table structures, long tables, lengthy texts, and multi-table contexts—challenges that have minimal impact on human performance \cite{mmqa2, multiTabQA}. Benchmarks such as HiTab \cite{hitab} and MULTIHIERTT \cite{multihiertt} have been instrumental in highlighting these challenges. HiTab features hierarchical multidimensional tables, while MULTIHIERTT further incorporates lengthy texts where answers may be embedded, as well as multi-table scenarios. Both benchmarks report model performances below 50\%, compared to a human accuracy of around 83\% on MULTIHIERTT. Similarly, benchmarks like MultiTableQA \cite{multiTabQA} and MMQA \cite{mmqa2}, which focus on multi-table question answering from well-structured databases such as those in the Spider benchmark, provide valuable insights into current model limitations. For instance, in MMQA the strongest model evaluated, o1-preview \cite{o1Preview}, achieves an exact match score slightly above 50\%, while human performance reaches approximately 89\%.

\paragraph{Scientific Document Understanding with Tables.} Scientific documents provide a rich test bed for information extraction and table extraction \cite{park2025can,sche2json,telin,axcell}. These papers typically contain complex ablation, analysis, and method‑comparison tables alongside extensive textual discussion, all of which demand sophisticated reasoning for accurate interpretation \cite{scienceBench,openscholars}. Building on this foundation, future work can harness scientific‑document data to develop higher‑level table‑reasoning systems that demand a broad repertoire of skills—such as trend detection, diagnostic assessment, and forecasting (see Figure \ref{fig:docUnderstanding}).

\subsection{Limited Generalization Across Tabular Representations}
Despite recent advances, current models still struggle to generalize across diverse tabular representations. Their performance on commonly used benchmarks can vary by up to 5\% depending on how closely input formats align with the data encountered during pretraining \cite{tableMeetsLLM}, as similarly observed by \citet{dailqsql} in the Text-to-SQL domain. 
Benchmarks highlight this issue by relying on a variety of input representations chosen based on convenience and accessibility. As demonstrated in our collection of major benchmarks (see Tables \ref{tab:tqa}, \ref{tab:table2text}, and \ref{tab:tfv}), tabular representations for the same type of task lack universality. Even when categorized under the same format, such as JSON, the internal structures can vary greatly \cite{feverous, hybridqa}, further complicating performance evaluations and introducing bias.

Efforts to address these inconsistencies are emerging. For example, \citet{TableQAKit} provides standardized serialization options such as Markdown and flattened text, though additional formats remain underexplored. Another line of research \cite{multimodalTabUnderstanding} focuses on visual representations of complex tables—such as Table Cell Locating and Merged Cell Detection—to generate serialized versions from images. Integrating these tasks into fine-tuning pipelines has proven beneficial.
\begin{figure}
    \centering
    \includegraphics[width=1\linewidth]{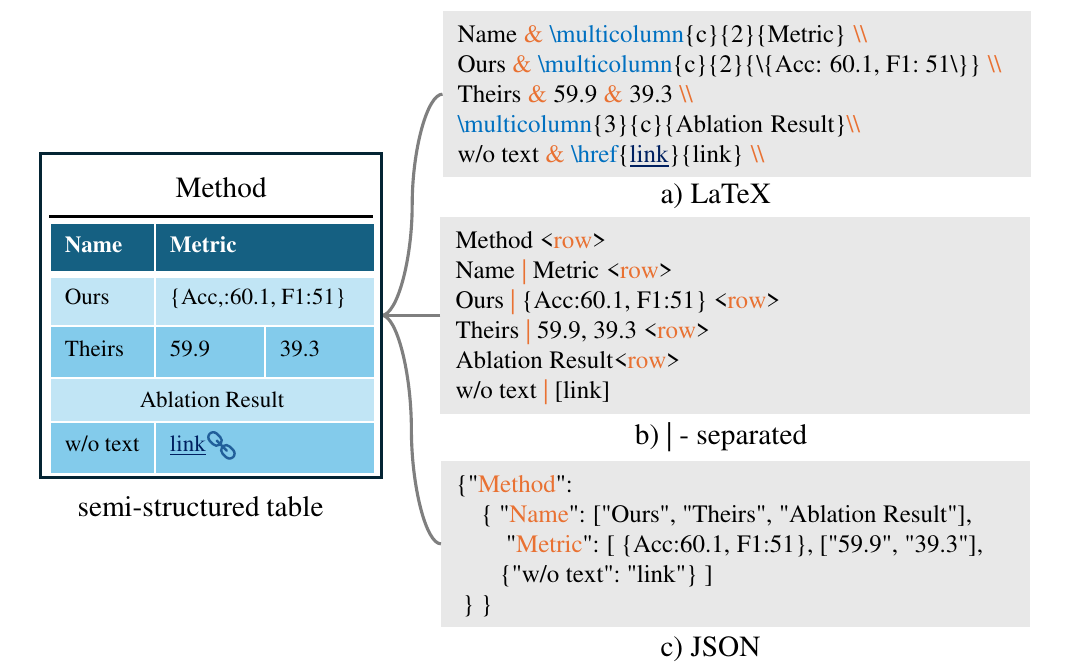}
    \caption{Comparison of serialization methods for semi-structured tables: a) LaTeX, b) X-separated, and c) JSON. Each method has its strengths and weaknesses in handling aspects such as nested value structures, row or column hierarchies, embedded document links, and flexible data types.}

    \vspace{-10pt}
    \label{fig:wellVsSemiStructure}
\end{figure}

Future research could explore serialization-to-serialization tasks, where models transform one format (e.g., JSON) into another (e.g., LaTeX or Markdown). Integrating such task could enhance models’ robustness to varied input styles and create opportunities for fine-tuning across multiple representations. Additionally, limited investigation has been conducted into the effectiveness of different representations for complex tables. For instance, LaTeX’s \texttt{\textbackslash multicolumn{}} command effectively captures hierarchical structures, whereas other formats may ignore this type of relationship during serialization process, as Figure \ref{fig:wellVsSemiStructure} shown.

\section{Modalities of Table Representation}
In this section, we introduce key tabular representations that are essential for enabling large models to process table data effectively. Since these models require one-dimensional input formats, structured, two-dimensional tables must be converted accordingly. This transformation, however, often results in the loss of valuable structural information. To address these challenges, various methods have been developed, including serialization, database schema representations, image-based formats, and specialized table encoders, as illustrated in Figure \ref{fig:inputRepresentation}. Recent studies \cite{tableMeetsLLM, actSql} demonstrate that model performance is sensitive to the chosen input representation, underscoring the data-dependent nature of current approaches to processing tabular data. Unfortunately, many existing benchmarks rely on representations selected primarily for convenience \cite{problemsInToTTo}, lacking of robust, unbiased comparisons.

\subsection{Serialization}

Serialization has long been a common method for representing tabular data, transforming tables into serialized text. Its primary advantages lie in compatibility with standard models and ease of access to existing formats, such as HTML or Markdown tables on the web, LaTeX tables in PDF documents, and JSON or key-value pairs in code environments (see Figure \ref{fig:inputRepresentation}). Most current benchmarks rely on serialization, as illustrated in Tables \ref{tab:tqa}, \ref{tab:table2text}, and \ref{tab:tfv}. Below, we highlight several noteworthy papers:

\paragraph{Sensitivity of Input Design.} Models are not only sensitive to different serialization formats, but variations in input design can also cause significant fluctuations in performance across table interaction tasks such as table partitioning, cell lookup, and reverse lookup \cite{tableMeetsLLM}. For example, omitting marked partitions or altering the input order has resulted in performance drops of up to 20\%, while removing example shots has led to accuracy deteriorations of as much as 50\%.

\paragraph{Sampling and Augmentation.} Long or multi-table inputs pose challenges for serialization due to model input length limitations, often resulting in truncation or data loss. To address these constraints, researchers have developed methods for sampling rows or columns that capture the key information in a table. Recent research \cite{tap4LLM} demonstrates that embedding-based sampling techniques, such as centroid and semantic-based sampling, outperform other approaches. Furthermore, they show a balanced combination of augmentation data (e.g., table sizes and keyword explanations) and sampled table text has proven effective in achieving better overall performance within token limits.


\subsection{Data Schema}
Another input representation for table is to provide the schema of tables rather than presenting the entire table content. Common schema representations include database structures in SQL and dataframes in pandas, as illustrated in Figure \ref{fig:inputRepresentation}. Using a data schema allows models to bypass input length limitations by focusing only on the structural blueprint of the data. However, this approach relies on strictly well-structured tables to be effective and loss of potential useful detailed content and value.

\paragraph{Sensitivity of Input Design.} Like serialization, models are not only sensitive to the schema format, but also its designs: \citet{actSql} evaluated schema input designs on GPT-3.5 and found that using three example rows yielded the best results. Additionally, they highlighted that model performance declines sharply when primary and foreign keys (PF keys) in the data schema are omitted, which \citet{isRetrievalSolved} also mentioned.

\paragraph{Normalized structure.}Given the trend toward schema-based methods and the improved results observed in Table QA tasks using Python or SQL code to interact with schema-based tables \cite{chainoftable, dinsql, dater}, exploring methods to convert complex data structures into more structured tables could be beneficial to enhance the compatibility of such methods.

\begin{table*}[ht]
\centering
\scriptsize
\begin{tabular}{l@{\hspace{2pt}}cc@{\hspace{2pt}}c@{\hspace{4pt}}c@{\hspace{4pt}}ccc}
\toprule
\footnotesize\textbf{Benchmark} & \footnotesize\textbf{Sources / Domain} &   \footnotesize\textbf{\# Q} & \footnotesize\textbf{\# T} & \footnotesize \textbf{Passage} & \footnotesize\textbf{Table Format} & \footnotesize\textbf{Output} & \footnotesize\textbf{Directions} \\ 
\midrule
WTQ \citeyearpar{WTQ} & Wikipedia & 22,033 & 2,108 & & HTML & cells & - \\ 
SQA \citeyearpar{SQA} & Wikipedia & 17,553 & 6,066 & & HTML & cells & Input Complexity\\ 
HybridQA \citeyearpar{hybridqa} & Wikipedia & 69,611 & 13,000 & \checkmark & JSON & text-span & Input Complexity\\ 
FetaQA \citeyearpar{fetaQA} & Wikipedia & - & 10,330 &  &Data Matrices & free-form & Answer Format \\ 
TAT-QA \citeyearpar{tatQA} & Financial Reports & 16,552 & 7,431 & \checkmark& Data Matrices & number & Domain, Input\\ 
OTT-QA \citeyearpar{ottQA}& Wikipedia & - & 45,841 & \checkmark &JSON & text-span & Input, Reasoning  \\
\midrule
AIT-QA \citeyearpar{aitQA} & Airline Industry & 515 & 113  && Data Matrices & cells & Domain, Input\\
FinQA \citeyearpar{finQA} & Financial Report & 8,281 & 2,789 & \checkmark &Data Matrices & number & Domain Knowledge\\
MMCoQA \citeyearpar{mmcoQA} & MMQA \citeyearpar{mmqa1} &  1,715 & 10,042 &\checkmark & JSON & text-span & Input Complexity\\
HiTab \citeyearpar{hitab} & Wikipedia, Statistic & 10,672 & 3,597 & & Row-Separated & text-span & Input Complexity\\
MULTIHIERTT \citeyearpar{multihiertt} & Financial Report & 10,440 &  2,513 & \checkmark & HTML & number & Input, Reasoning\\
\midrule
Open-WikiTable \citeyearpar{openWikiTable} & Wikipedia & 67,023 & 24,680 & & Row-Separated & text-span, SQL & Answer Format\\

QTSUMM \citeyearpar{qtsumm} & Wikipedia & 7,111 & 2,934 & &Data Matrices & free-form & Answer Format\\
TEMPTABQA \citeyearpar{tempTABQA} & Wikipedia & 11,454 & 1,208 & & JSON, HTML & text-span &  Reasoning Difficulty\\
CRT-QA \citeyearpar{crtQA} & TabFact \citeyearpar{tabFact} & 1,000 & 423 & & Row-Separated & text-span & Reasoning Difficulty\\
IM-TQA \citeyearpar{imTQA} &Baidu Encyclopedia &5,000 & 1,200 & & Index Mapping & text-span & Input Complexity\\
TabCQA \citeyearpar{TabCQA} & Financial Report & 109,089 & 7,041 & & \makecell{Text Template, \\Value Pair} & text-span & Input Complexity\\
MultiTabQA \citeyearpar{multiTabQA} & \makecell{Spider \citeyearpar{spider}, Synthetic, \\TAPEX \citeyearpar{tapex} Corpus} & 136,461 & - & & Row-Separated & sub-table & Answer, Input\\

TABMWP \citeyearpar{tabMWP} & Online Learning Web& 38,431 & 37,544 & & \makecell{Row-Seperated, \\SpreadSheet, Image} & free-form & Reasoning Difficulty\\

\midrule
FREB-TQA \citeyearpar{frebTQA} & \makecell{WTQ, WikiSQL \citeyearpar{wikiSQL}, \\ SQA, TAT-QA} & 75,205 & 8,590 & &Data Matrices & text-span & Input, Reasoning\\
Text2Analysis \citeyearpar{text2analysis} & Data Analysis Libraries & 2,249 & 347 & & - & code, text & Reasoning Difficulty\\

MMQA \citeyearpar{mmqa2} & Spider \citeyearpar{spider} & 3,313 & 3,312 & &JSON & sub-table & Input Complexity\\
\bottomrule
\end{tabular}

\caption{Summary of benchmarks for Table-based Question Answering. \textbf{Sizes} shows the number of questions and tables. \textbf{Passage} indicates if an input passage is included. \textbf{Directions} categories each benchmark’s primary focus compare to previous ones.}
\label{tab:tqa}
\end{table*}
\subsection{Image}

With the advancement of MLLMs, there is growing interest in using images as an input format due to their adaptability, accessibility, and ability to preserve structural information \cite{vistabnet}. Specifically, \citet{multimodalTabUnderstanding} achieved superior results using images with a fine-tuned LLaVA model \cite{llava}, outperforming models with OCR and serialization settings. They found that additional training focused on table structure understanding—such as cell extraction and cell location—enhance the model's ability to accurately interpret tables.

\paragraph{Image resolution.}While images offer the advantage of preserving the original table layout, they face constraints similar to serialization: the amount of data they can present is limited by image size and resolution, which can significantly impact model performance \cite{ImgResolutionIsImportant}. As tables grow larger, the information becomes blurred at a fixed resolution, leading to deteriorated performance. One potential approach is to use images as supplemental input alongside serialized text or data schema \cite{unifyingTextTablesandImages}. This combined input strategy could potentially allow the model to receive structural information directly from the image while accessing detailed content from the text-based format. However, to the best of our knowledge, systematic evaluations of this approach remain lacking.

\subsection{Table Encoder}

Specific table encoder designs have been employed in smaller-scale language models to handle table-related tasks, utilizing various embeddings such as column-based \cite{tabbie}, row-based \cite{tapas}, tree-structured \cite{tuta}, and graph-based embeddings \cite{gtr}. Building on these approaches, recent work has demonstrated a trend toward employing specialized encoders in larger base models, effectively creating table foundation models \cite{tableFoundationModel, tablegpt2, tabdpt}. In particular, TableGPT2 leverages a specialized table encoder—with column- and row-wise attention—to integrate tabular data during the pretraining and fine-tuning stages of 7B and 72B base models \cite{tablegpt2}, outperforming other table generalist models across a range of tasks while remaining competitive with task-specific methods.

\section{Table-Related Tasks}

In this section, we introduce key table-related tasks such as Table Question Answering (TQA), Table-to-Text, and Table Fact Verification (TFV), along with other intriguing applications like leaderboard construction that actively utilize tables.

\begin{table*}[ht]
\centering
\scriptsize

\begin{tabular}{lcc@{\hspace{2pt}}ccccc}
\toprule
\footnotesize\textbf{Benchmark} & \footnotesize\textbf{Sources / Domain} & \footnotesize\textbf{\# Q} & \footnotesize\textbf{\# T} & \footnotesize\textbf{Table Format} & \footnotesize\textbf{Focus} & \footnotesize\textbf{Text Input} & \footnotesize\textbf{Directions} \\ 
\midrule
Rotowire \citeyearpar{rotowire} & NBA & - & 4,853 & JSON & N/A & & Domain Knowledge\\ 
ToTTo \citeyearpar{totto} & Wikipedia & 134,161 & 83,141 & Index Mapping & Highlight Span & Caption & -\\ 
Logic2Text \citeyearpar{logic2text} & WikiTable & 10,800 & 5,600 & Row-Separated & N/A  & & Logic Summarization\\ 
LogicNLG \citeyearpar{logicalNLG} & TabFact \citeyearpar{tabFact} & 37,000 & 7,300 & Data Matrices & N/A  & &Logic Comparison\\ 
\midrule

SciGen \citeyearpar{sciGen} & Scientific Paper & 53,000 & - & Row-Separated & N/A & Caption & Domain Knowledge\\
NumericNLG \citeyearpar{numericNLG} & Scientific Paper &  1,300 & 1,300 & JSON & N/A  & Caption & Domain Knowledge\\
FetaQA \citeyearpar{fetaQA} & ToTTo \citeyearpar{totto} & - & 10,330 & Matrices & Text Query &  & Input Complexity\\ 
&E2E \citeyearpar{e2e}, WTQ&&&&&&\\
DART \citeyearpar{dart} & WikiTable \citeyearpar{openWikiTable} & 82,191 & 5,623 & XML, JSON &N/A  &  Table Title& Table Structure\\ 
&WebNLG \citeyearpar{webNLG} &&&&&&\\
\midrule
QTSUMM \citeyearpar{qtsumm} & Wikipedia & 7,111 & 2,934 & Data Matrices & Text Query & & Input Complexity\\
FindSUM \citeyearpar{findSUM} & Company Report & - & 21,125 & Data Matrices & N/A & Long Text & Input Complexity\\

\bottomrule
\end{tabular}

\caption{Summary of benchmarks for Table-to-Text and Table Summarization. \textbf{Focus} specifies the subset of table content intended for natural language generation, while N/A indicates the entire table should be transformed to natural language.}
\label{tab:table2text}

\end{table*}

\begin{table*}[ht]
\centering
\scriptsize

\begin{tabular}{l@{\hspace{2pt}}c@{\hspace{2pt}}c@{\hspace{2pt}}cccc}
\toprule
\footnotesize\textbf{Benchmark} & \footnotesize\textbf{Sources / Domain} &  \footnotesize\textbf{\# Q} & \footnotesize\textbf{\# T} & \footnotesize\textbf{Table Format} & \footnotesize\textbf{Output}  & \footnotesize\textbf{Directions} \\ 
\midrule
TabFact \citeyearpar{tabFact} & Wikipedia & 117,843 & 18,000 & Row-Separated & S, R & - \\ 
InfoTabs \citeyearpar{infotabs} & Wikipedia & 23,738 & 2,540 & HTML, JSON & S, R, N & Output Format\\
FEVEROUS \citeyearpar{feverous} & Wikipedia & 87,062 & - & JSON / Mapping &S, R, N & Output Format\\ 
SEM-TAB-FACTS \citeyearpar{semTabFacts} & Science & 5,715 & 2,961 & XML & S, R, N, EC & Domain Knowledge\\ 
XInfoTabs \citeyearpar{xinfotabs} & InfoTabs & 23,738 & 2,540 & JSON & S, R, N & Multi-Language\\
EI-InfoTabs \citeyearpar{eiINFOTABS} & InfoTabs & 23,738 & 2,540 & JSON & S, R, N & Indic-Language\\
SciTab \citeyearpar{scitab} & SciGen\cite{sciGen} & 1,255 & - & JSON / Mapping& S, R, N & Domain Knowledge\\ 
\bottomrule
\end{tabular}

\caption{Summary of benchmarks for Table-based Fact Verification. \textit{S} in the output denotes Supported, \textit{R} represents Refuted, \textit{N} stands for Neither or Not Enough Evidence, and \textit{EC} refers to Evidence Cells.}

\label{tab:tfv}
\end{table*}

\subsection{Table Question Answering}
TQA\footnotemark is one of the most common and well-studied table tasks, with various benchmarks developed as shown in Table \ref{tab:tqa}. It typically involves a free-form question and a single table, sometimes accompanied by an optional passage or passage links, and the output is expected to be information derived from the table or passage, generally presented as cell spans, calculated values, or minimal text spans.

TQA benchmarks have expanded significantly over the past two years, inspiring future work across multiple directions, including domain knowledge, answer format, input complexity, and reasoning difficulty. Domain-specific benchmarks now better reflect real-world scenarios in fields such as airlines \cite{aitQA} and finance \cite{tatQA, finQA}. Answer formats have also diversified, with benchmarks requiring free-form responses \cite{fetaQA, qtsumm, revisitingLFTQA} and SQL queries \cite{openWikiTable}, beyond traditional cell values or text spans. Input complexity has increased through multi-table datasets \cite{multiTabQA, multihiertt}, hierarchical tables \cite{hitab}, and semi-structured tables \cite{tabMWP}, which challenge models to navigate intricate structures. Reasoning requirements have similarly intensified, incorporating hypothetical questions \cite{hypotheticalQA}, implicit time-based inference \cite{tempTABQA}, and sequential or conversational queries \cite{SQA, mmcoQA, TabCQA}. Overall,  recent benchmarks generally demand more complex reasoning steps and operations to yield accurate answers\footnotetext{For a more comprehensive understanding of TQA, see this curated list of relevant papers: \url{https://github.com/lfy79001/Awesome-Table-QA}}.

\subsection{Table-to-Text and Table Summarization}
Table-to-Text and Table Summarization are table tasks initially developed to evaluate whether models could accurately interpret and describe table content. In these tasks, the input typically includes a table, sometimes with specified cell spans as shown in the \textit{Focus} column in Table \ref{tab:table2text}. If a span or region is provided, the model generates a textual description or summary of that specific area; if not, it summarizes the entire table. With advances in models’ table understanding, this task has become less prominent, as the number of related publications has steadily decreased since 2021.
\vspace{-5pt}
\paragraph{Query Focused Summarization.}A recent, noteworthy benchmark in this area is QTSUMM \cite{qtsumm}, which requires models to generate text-based summaries of specific table regions in response to questions. By integrating the aspect of table search based on textual queries from TQA with the descriptive demands of Table-to-Text, QTSUMM introduces new complexities that push models to move beyond simple fact retrieval. Notably, QTSUMM includes “why” questions, prompting models to reason about underlying causes or explanations—a shift that aligns more closely with human interests and highlights the importance of generating responses that incorporate causal understanding and contextual depth.
\vspace{-5pt}
\paragraph{Lack of Multilingual Benchmarks.}A notable gap in current research is the absence of multilingual benchmarks for table-to-text tasks. As highlighted in \cite{surveydatatotextnlg}, to the best of our knowledge, no table-to-text benchmarks exist in languages other than English, significantly limiting the applicability and inclusivity of this task.

\begin{figure}
    \centering
    \includegraphics[width=1\linewidth]{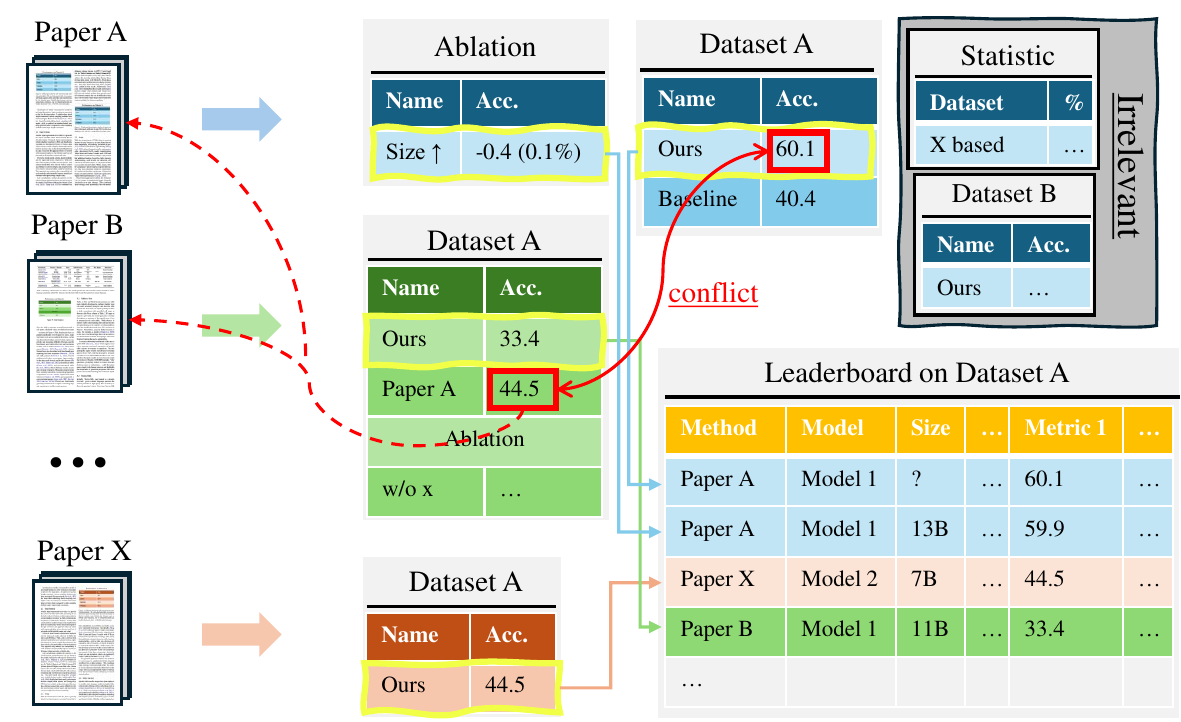}
    \caption{Illustration of automatic leaderboard construction pipeline. Results are extracted from ablation and performance tables in each paper. The \textcolor{red}{red} line highlights inconsistency across paper that may require examination across texts.}
    \vspace{-5pt}
    \label{fig:leaderboard}
\end{figure}

\subsection{Table Fact Verification}
Table Fact Verification (also referred to as Table Reasoning or Table Natural Language Inference) is a task designed to assess fact-searching and logic inference capabilities within tables. In this task, the input typically consists of a statement or claim alongside a reference table. The model's output is a verification label—such as “Supported,” “Refuted,” or “Not Enough Information”—indicating whether the claim aligns with the table content. Some benchmarks also require a justification for the answer, as shown in Table \ref{tab:tfv}. Recent methods have enabled models to achieve over 80\% accuracy on widely used benchmarks like TabFact and FEVEROUS \cite{tableMeetsLLM, dater, chainoftable}, demonstrating substantial progress in fact-checking within tabular data. However, scenarios involving longer contexts, multiple tables, or complex table structures remain unassessed.

\subsection{Leaderboard Construction}
Beyond the widely studied tasks, an intriguing direction proposed by \citet{axcell} is leaderboard construction. This task aims to streamline the comparison of experimental results within a research domain through scientific papers, offering a concise and structured view of progress. 

Existing methods, such as those proposed in \cite{axcell, telin}, have made notable strides in automating this process. These approaches typically employ pipelines that classify and extract data from performance and ablation tables in scientific papers, leveraging techniques like Named Entity Recognition (NER) or string matching to form tuples (Task, Dataset, Metric) or quadruples (Task, Dataset, Metric, Score). Such methods provide a foundational framework for building leaderboards and have proven effective in capturing basic performance comparisons across different methods and datasets. However, as scientific tasks and methodologies grow increasingly complex, these pipelines face limitations. Tasks often require varying schemas to account for unique aspects, and surface-level extraction may not fully capture the nuances of more intricate experiments or analyses. For instance, discrepancies in reported results between papers, as illustrated in Figure \ref{fig:leaderboard}, often necessitate a deeper comparison and reasoning over both tables and textual content to resolve.

\subsection{Other Tasks}
Emerging new table-related tasks include innovations such as tabular synchronization across languages \cite{infosync} and column name abbreviation expansion \cite{nameguess}. Among these, Text-to-Table has gained increasing attention in 2024 \cite{textToTableBad, textToTableKG, textToTableTuple}. The task was first formalized by \citet{textToTable} as a sequence-to-sequence task by inversely applying table-to-text datasets. Recent studies have explored various methods, such as incorporating knowledge graphs \cite{textToTableKG}, to enhance its utility as a data integration task for field like finance, medicine, and law.

\section{Further Reading}
For readers seeking deeper insights into table-related research areas, several survey papers offer valuable perspectives. For methodologies aimed at improving table reasoning with LLMs, work by \citet{survey2} provides a detailed taxonomy and an analysis of emerging trends. \citet{survey3} explores prompting and training techniques for table-related tasks in the context of LLMs and VLMs. Meanwhile, \citet{survey4, survey5} presents a focused analysis of transformer-based, smaller-scale models designed for tabular data. For an in-depth perspective, the comprehensive 30-page survey by \citet{survey1} provides an extensive overview of table understanding tasks, datasets, and corresponding fundamental methods.




\section*{Limitations}
This study presents a comprehensive survey of table-related tasks with LLMs and MLLMs, highlighting key trends and emerging opportunities. While we have made our best effort to provide a thorough review, certain limitations remain. Due to space constraints, we focus on summarizing the main trends rather than providing exhaustive technical details for each approach. Our selection of works primarily draws from major NLP conferences, including ACL, EMNLP, NAACL, and ICLR, along with relevant studies from other domains and preprints. While we strive to incorporate the latest research, many new works continue to emerge during our submission of this paper. Given the rapid evolution of this field, our survey offers a snapshot of current progress rather than a definitive account. We will continue to track developments and refine our analysis in future updates.

\bibliography{custom}

\appendix

\begin{table*}[ht]
\centering
\scriptsize

\begin{tabular}{lccccc}
\toprule
\footnotesize\textbf{Benchmark} & \footnotesize\textbf{Sources / Domain} & \footnotesize\textbf{Sizes} & \footnotesize\textbf{Input Format} & \footnotesize\textbf{T / Q} & \footnotesize\textbf{Directions} \\ 
\midrule
WikiSQL \citeyearpar{wikiSQL} & Wikipedia & 80,654 & Row Header, Row-Separated  & 1.0 & -\\ 
Spider \citeyearpar{spider} & Academic Databases, Online CSV, WikiSQL & 10,181 & Table(col), Type, PF & 1.6 & - \\ 
\midrule

SEDE \citeyearpar{sede} & Stack Exchange & 12,023 & Table(col), Type, PF & 1.3 & Noise Utterance\\ 
SpiderDK \citeyearpar{spiderDK} & Spider & 535 & Table(col), Type, PF & > 1 & Domain Knowledge\\ 
SpiderSyn \citeyearpar{spiderSYN} & Spider & 8,034 & Table(col), Type, PF & > 1 & Query Perturbation\\
SpiderRealistic \citeyearpar{spiderRealistic} & Spider & 508 & Table(col), Type, PF &  > 1 & Query Perturbation\\
MIMICSQL \citeyearpar{mimicSQL} & Electronic Medical Records & 10,000 & Row Header, Row-Separated & 1.8 & Domain Knowledge\\
KaggleDBQA \citeyearpar{kaggledbqa} & \makecell{ATIS, GeoQuery, Restaurants, Yelp,\\ Academic, IMDB, Scholar, Advising} & 272 & Table(col), Type, PF, context & 1.2 & Domain Knowledge\\ 
\midrule
ADVETA \citeyearpar{adveta} & Spider, WikiSQL, WTQ & - & Table(col), Type, PF & > 1 & Table Perturbation\\ 
BIRD \citeyearpar{bird} & Kaggle, Machine Learning platform & 12,751 & Table(col), Type, PF, context & > 1 & Table Size\\
Dr.Spider \citeyearpar{drSpider} & Spider & 15,000 & Table(col), Type, PF & > 1 & Table, Query Perturbation\\
EHRSQL \citeyearpar{ehrsql} & Electronic Medical Records & 24,000 & Table(col), Type, PF & 2.4 & Domain, Reasoning\\

ScienceBench \citeyearpar{scienceBench} & CORDIS, SDSS, OncoMX & 6,000 & Table(col), Type, PF & > 1 & Data Synthesis, Domain\\

\midrule
TrustSQL \citeyearpar{trustsQL} & ATIS, Advising, EHRSQL, Spider & 27,784 & CREATE(EoT) & > 1 & Reasoning\\
Spider2 \citeyearpar{spider2} & Cloud Data Warehouses & 632 & Table(col), PF& > 1 & Reasoning, Table Size\\ 
Spider2V \citeyearpar{spider2v} & Cloud Data Warehouses & 494 & Agent Workspace & > 1 & Input Modality\\ 

\bottomrule
\end{tabular}

\caption{Summary of benchmarks for Text-to-SQL. \textbf{Sizes} refers to the number of SQL query pairs, and \textbf{T/Q} indicates the number of tables required to answer a single query.}

\label{tab:sql}
\end{table*}
\section{Text-to-SQL}
\label{sec:text2sql}
Text-to-SQL is a semantic parsing task that is highly relevant to table-based applications: given a natural language question, the model must generate a SQL query that accurately captures the intent of the query. Over time, these tasks have evolved to incorporate additional contextual information—such as table schemas and optional sample rows—with the evaluation focus shifting from exact match (EM) to execution accuracy (EX) as the primary metric. A prominent benchmark in this area, Spider \cite{spider}, significantly increased task complexity by introducing databases composed of multiple tables, foreign keys, and the requirement to employ a variety of functions.

Building on Spider, several adaptations and extensions have broadened the task’s scope and complexity. Multilingual adaptations \cite{cSpider, vietSpider, multiSpider} expanded Text-to-SQL to cross-lingual and multilingual settings, enabling SQL generation across diverse languages. Other extensions include Spider-DK \cite{spiderDK}, which incorporates domain knowledge, and Spider-Syn \cite{spiderSYN} and Spider-Realistic \cite{spiderRealistic}, which obscure schema-related words or column names to simulate noisy utterances and more realistic queries.

Text-to-SQL has been well-studied with question decomposition pipelines \cite{dailqsql, dater, chainoftable}, with current models nearing saturation on some commonly used benchmarks.

\paragraph{Effect of Noisy Input.}Beyond evaluation issues, Text-to-SQL faces inherent challenges, especially when handling ambiguity, or on very large tables. As noted in \cite{isRetrievalSolved}, performance drops significantly without PF keys, as variations in column names across tables and limited sample rows complicate element matching. Moreover, as highlighted in \cite{spider2, deathOfSchemaLink}, model performance deteriorates sharply when processing extremely long database schema, a scenario prevalent in real-world industrial databases.

\section{Responsible NLP Miscellanea}
\subsection{AI Assistants}

We acknowledge the use of GPT-4o and GPT-o3-mini for grammar checking and word polishing.

\end{document}